%% file: main.tex
\documentclass[10pt,twocolumn,letterpaper]{article}

\usepackage{iccv}              
\usepackage{threeparttable}
\usepackage{multirow}
\usepackage{natbib}
\usepackage{xcolor}
\usepackage{colortbl,booktabs}

\definecolor{iccvblue}{rgb}{0.21,0.49,0.74}
\usepackage[pagebackref,breaklinks,colorlinks,allcolors=iccvblue]{hyperref}

\def\paperID{*****} 
\def\confName{ICCV}
\def\confYear{2025}

\title{USIS16K: High-Quality Dataset for Underwater Salient Instance Segmentation}

\author{Lin Hong, Xin Wang$^{*}$, Yihao Li, Xia Wang\\
Harbin Institute of Technology (Shenzhen)\\
{\tt\small Corresponding: wangxinsz@hit.edu.cn}}

\begin{document}
\maketitle
\input{0_abstract}

\input{1_intro}
\input{2_dataset}
\input{3_benchmark}
\input{4_conclusion}
\input{acknowledge}
{
    \small
    \bibliographystyle{plain} 
    \bibliography{main} 
}

\end{document}

%% file: 0_abstract.tex
\begin{abstract}
Inspired by the biological visual system that selectively allocates attention to efficiently identify salient objects or regions, underwater salient instance segmentation (USIS) aims to jointly address the problems of \textbf{where to look} (saliency prediction) and \textbf{what is there} (instance segmentation) in underwater scenarios.
However, USIS remains an underexplored challenge due to the inaccessibility and dynamic nature of underwater environments, as well as the scarcity of large-scale, high-quality annotated datasets. 
In this paper, we introduce \textbf{USIS16K}, a large-scale dataset comprising 16,151 high-resolution underwater images collected from diverse environmental settings and covering 158 categories of underwater objects. 
Each image is annotated with high-quality instance-level salient object masks,
representing a significant advance in terms of diversity, complexity, and scalability.  
Furthermore, we provide benchmark evaluations on underwater object detection and USIS tasks using USIS16K. 
To facilitate future research in this domain, the dataset and benchmark models are publicly available at:~\href{red}{https://github.com/LinHong-HIT/USIS16K}.
\end{abstract}

%% file: 1_intro.tex
\section{Introduction}
\label{sec:intro}
Inspired by the biological visual system, salient instance segmentation (SIS)~\cite{li2017instance}, refers to the ability to identify individual object instances within the detected salient regions of an observed scene, has become a well-established problem in computer vision and robotics. SIS has been modeled through the combination of salient object detection~\cite{redmon2016you} and instance segmentation~\cite{he2017mask}, finding widespread applications in autonomous navigation~\cite{li2021ion} and robotic manipulations~\cite{xie2021unseen}. 
Over the past few years, research on SIS has made significant strides within terrestrial environments~\cite{pei2022transformer,liu2021scg,10614124}, leading to the development of many practical models that have been deployed for various real-world tasks and applications. 
This progress has been largely driven by the availability of large-scale, well-annotated datasets~\cite{li2017instance,fan2018salient,pei2022transformer,10614124}, advances in computing hardware, and the emergence of powerful deep learning architectures.
More recently, the growing demands of marine ecological monitoring, underwater resource exploration, and underwater robotic operations have given rise to a specialized branch of SIS--underwater salient instance segmentation (USIS). 
USIS focuses on identifying individual object instances within the detected salient regions of an underwater image, which has attracted increasing research interest worldwide~\cite{lian2024diving}.

However, despite the growing shift from terrestrial to underwater environments and increasing research interest in USIS, the field remains in its early stages of development. A primary challenge stems from the fundamental differences between visual content captured in underwater environments and that from terrestrial domains. These differences arise from underwater-specific object categories, dynamic waterbody patterns, unavoidable light attenuation, and optical distortion artifacts~\cite{akkaynak2018revised}. As a result, SIS methods developed for terrestrial scenarios cannot be directly applied to underwater images without substantial retraining or fine-tuning.
An overview of existing datasets related to USIS is provided in Table~\ref{tab:dataset_review}. While several datasets~\cite{9340821,lian2023watermask,lian2024diving,zheng2024coralscop} have been introduced to support USIS research, most lack eye-tracking data as ground truth for accurately identifying salient objects or regions during dataset construction. Furthermore, they do not offer the aligned ground truth required for underwater salient object detection and underwater instance segmentation. In addition, these datasets fail to capture the diversity and complexity of real-world underwater environments, which limits their effectiveness for training USIS models with strong generalization capabilities.
 
\begin{table}[ht]
\begin{center}
\caption{Statistics of existing datasets related to USIS. Cate.: number of object categories, BB: bounding boxes, PW: pixel-wise masks. Det.: object detection, Seg.: object segmentation, Ins.: instance segmentation, Sem.: semantic segmentation, USOD: underwater salient object detection.}
\resizebox{\linewidth}{!}{ 
\begin{tabular}{l|c|lc|cc|ccccc}
\toprule
\rowcolor[gray]{0.7} Datasets  &Year &Size &Cate. &BB &PW  & Det. & Seg. & Ins. &Sem. & USOD\\
\hline
SUIM~\cite{9340821}       &2020  &1635  &8    &   &  \checkmark &  & & & & \checkmark \\
\rowcolor[gray]{0.9} USOD~\cite{Islam-RSS-22} &2022  &  300    & -  &  & \checkmark &   & & & & \checkmark \\
LIACI~\cite{9998080}      &2023  &1893  &10   &   & \checkmark   &  &  & & \checkmark \\
\rowcolor[gray]{0.9} USOD10K~\cite{10102831}    &2023  &  10,255    & 70  &  & \checkmark   & & & & & \checkmark \\

UIIS~\cite{lian2023watermask}    &2023  &  4628    & 7  & & \checkmark   &  & & \checkmark \\

\rowcolor[gray]{0.9} USIS10K~\cite{lian2024diving}    &2024  &  10,632    & 7  & & \checkmark  &  & & \checkmark & &\checkmark \\

CoralMask~\cite{zheng2024coralscop}   &2024  &41,297   &1 & & \checkmark   & &\checkmark &\\

\rowcolor[gray]{0.9} UIIS10K~\cite{li2025uwsam}   &2025  & 10,048   & 10  &  & \checkmark   &  &  &  \checkmark & &  \\
\hline
\textbf{{USIS16K (ours)}}        &2025  &16,151   &158   &\checkmark &\checkmark  &\checkmark & &\checkmark & &\checkmark \\
\bottomrule
\end{tabular}}
\label{tab:dataset_review}
\end{center}
\end{table}

To address the challenges and advance the research progress of the USIS, we hold the view that large-scale datasets and advanced models are interconnected and mutually reinforcing. To this end, this work provides a large-scale, high-quality dataset named USIS16K for the research community. USIS16K consists of 16,151 underwater images spanning 158 categories of commonly encountered underwater objects. Based on this dataset, we conduct benchmark evaluations on both underwater object detection and USIS tasks, providing baselines to support future research.

%% file: 2_dataset.tex
\begin{figure*}[t]
\begin{center}
\includegraphics[width=1\linewidth]{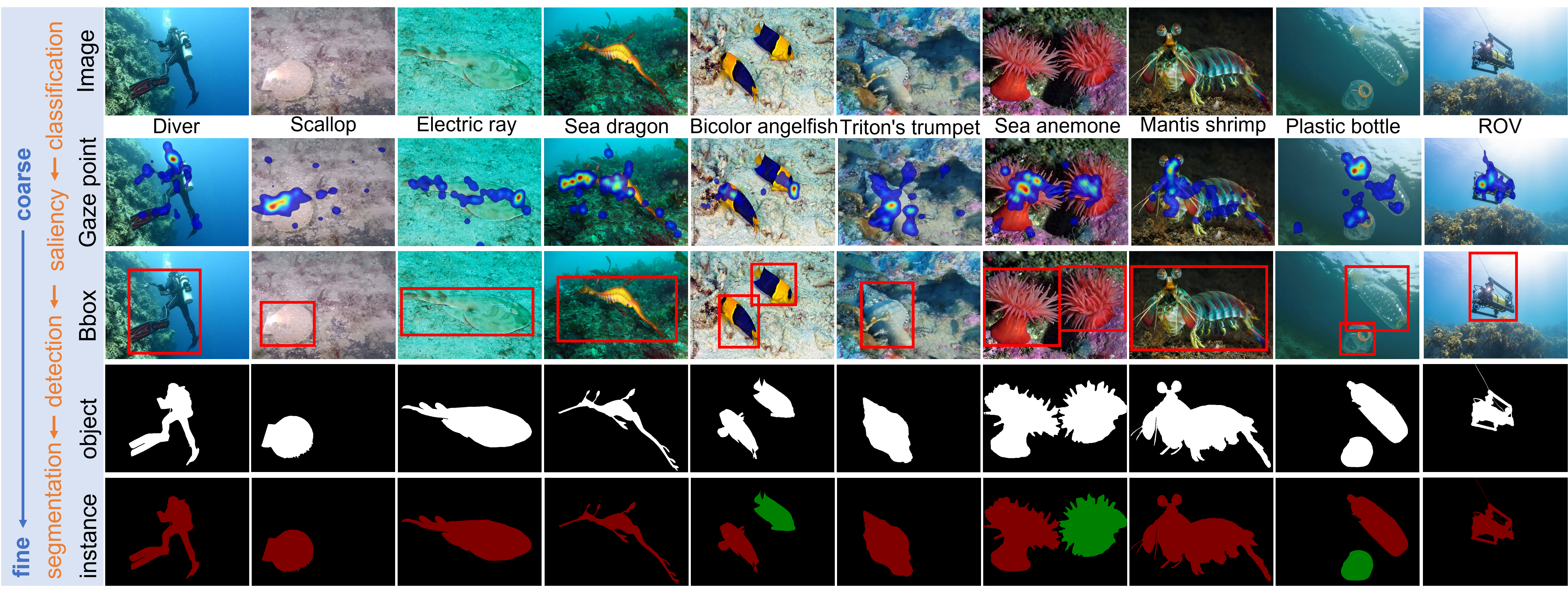}
\end{center}
   \caption{Underwater images in the USIS16K dataset are annotated with multi-level labels. From top to bottom: raw image with classification, gaze points, bounding boxes, pixel-wise salient object masks, and instance-level masks.}
\label{fig:dataset}
\end{figure*}

\section{USIS16K dataset}
\label{sec:USIS16k}
The emergency of large-scale datasets~\cite{li2017instance,fan2018salient,10614124} has fostered many effective SIS models within the terrestrial field. With this in mind, to advance the USIS research, this paper introduces a new large-scale dataset, named USIS16K. Example samples for a quick view of the USIS16K dataset are shown in Figure~\ref{fig:dataset}. 

\subsection{Dataset Construction}
The construction of the USIS16K dataset involves four sequential steps: 1) image collection, 2) image filtering, 3) image annotation, and 4) dataset splitting.
\subsubsection{Image Collection}
A large proportion of the images in the USIS16K dataset is sourced from the internet, aligning with the common practice of utilizing internet available images in the construction of datasets such as COCO~\cite{lin2014microsoft} and ImageNet~\cite{deng2009imagenet}. For image collection, the initial step involves the search for images and videos containing common underwater objects on popular search engines such as Google, Bing, and Flickr. 
In addition to internet sources, representative underwater images from the USIS16K dataset~\cite{10102831} are incorporated to enrich the dataset. Moreover, a significant number of in-situ underwater images were captured during underwater engineering tasks performed by robotic systems under different lighting conditions. As a result, over 35,000 candidate underwater images have been collected to form the dataset.
\subsubsection{Image Filtering}
Three volunteers are organized to manually filter the candidate images. Prior to the filtering process, they participated in three training sessions to ensure consistency and quality: (1) understanding the common sense of salient object detection and instance segmentation, (2) familiarizing themselves with the construction processes of existing large-scale SIS datasets in the terrestrial domain, and (3) comprehending the overarching goal of constructing the USIS16K dataset. During the image filtering stage, the volunteers reviewed the candidate images, excluding those deemed unsatisfactory and removing duplicates or damaged images based on their training. After this process, a total of 16,151 images are retained for professional annotation.

\subsubsection{Image Annotation}
\begin{figure}[ht]
\begin{center}
\includegraphics[width=0.99\linewidth]{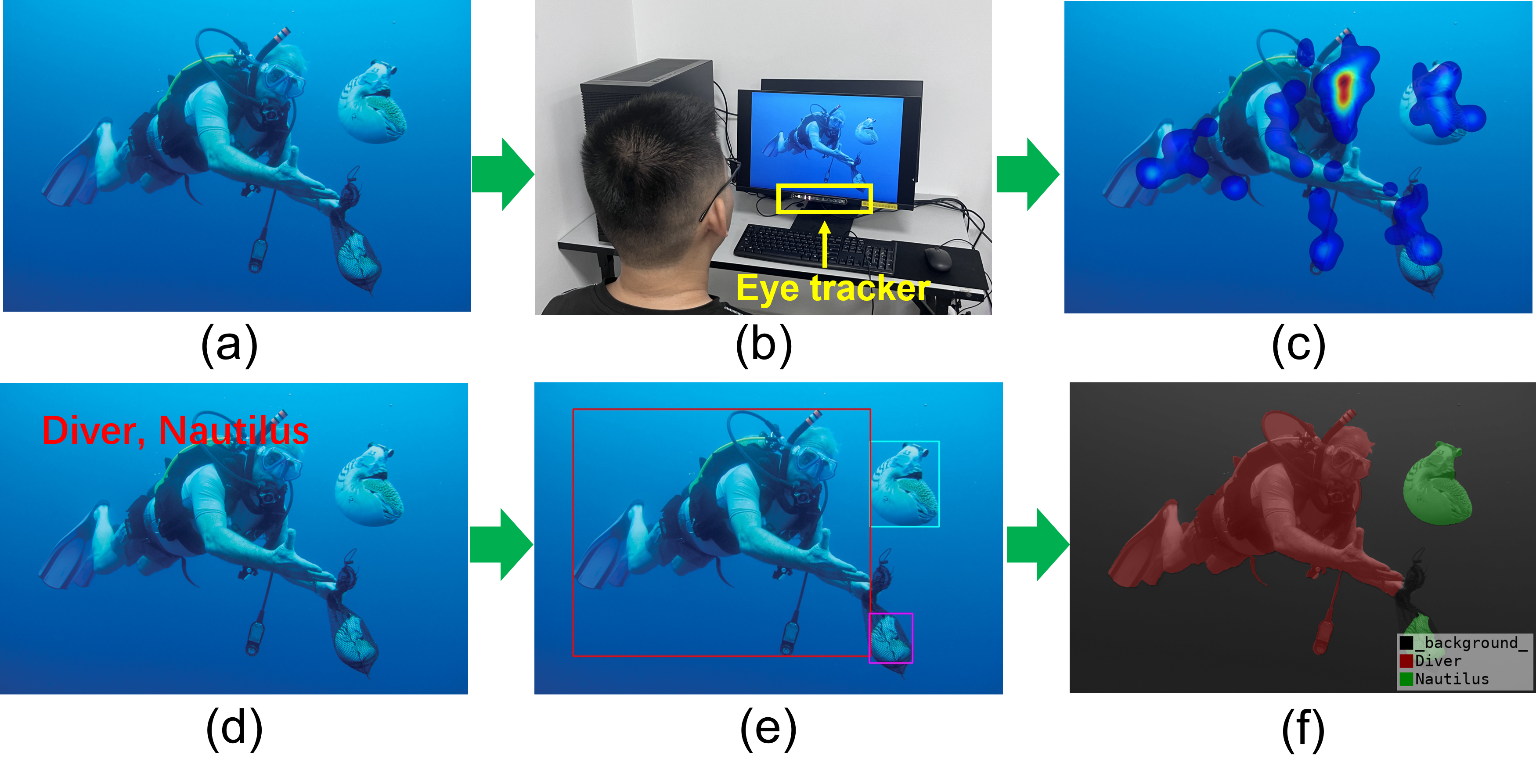}
\end{center}
   \caption{Image annotation pipeline: (a) underwater image, (b) gaze point prediction using an eye tracker, (c) salient object determination, (d) salient object classification, (e) generation of bounding boxes for each salient object, and (f) generation of instance-level masks for each salient object.}
\label{fig:annotation_pipeline}
\end{figure}

Accurate and high-quality annotations are essential for constructing the USIS16K dataset. The image annotation pipeline is illustrated in Figure~\ref{fig:annotation_pipeline}. A team of ten trained annotators was organized and provided with comprehensive instructions to carry out multi-level annotations. The process began with identifying salient objects or regions in candidate images, guided by eye-tracking data. These salient objects were then subjected to object classification.
Subsequent annotation proceeded systematically from coarse to fine granularity, including bounding box labeling, salient object segmentation, and instance-level mask generation. This rigorous annotation workflow resulted in the USIS16K dataset, which comprises 16,151 underwater images with comprehensive and consistent multi-level annotations.

\subsubsection{Dataset Splitting}
To ensure reliable training, validation, and testing of deep learning models on the USIS16K, it is essential to maintain an adequate and balanced distribution of samples for each object category across the dataset splits. Following common practice, the USIS16K employs a 7:2:1 split ratio at the category level, yielding 11,306 samples for the training set, 3,230 for the validation set, and 1,615 for the test set.
\subsection{Dataset Diversity}
\subsubsection{Taxonomic System}
\begin{figure}[t]
\begin{center}
\includegraphics[width=0.9\linewidth]{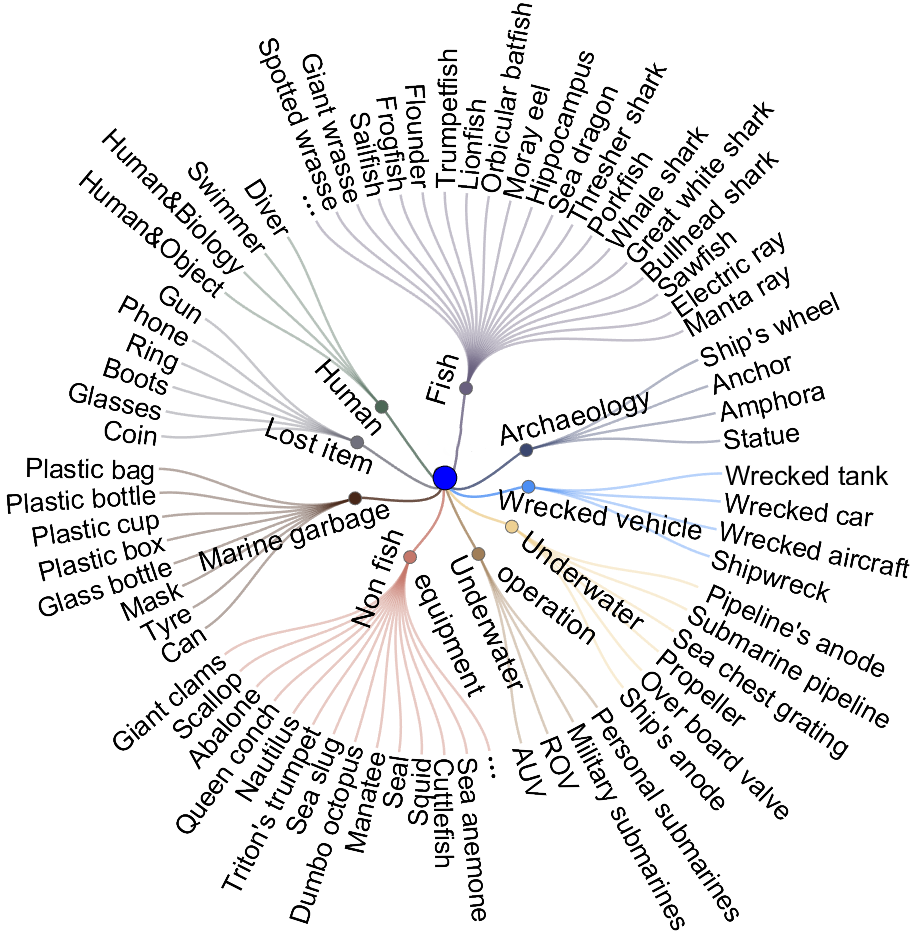}
\end{center}
   \caption{Hierarchical taxonomic system of USIS16K dataset. Best viewed on screen and zoomed-in for details.}
\label{fig:tax_sys}
\end{figure}

\begin{figure}[t]
\begin{center}
\includegraphics[width=0.99\linewidth]{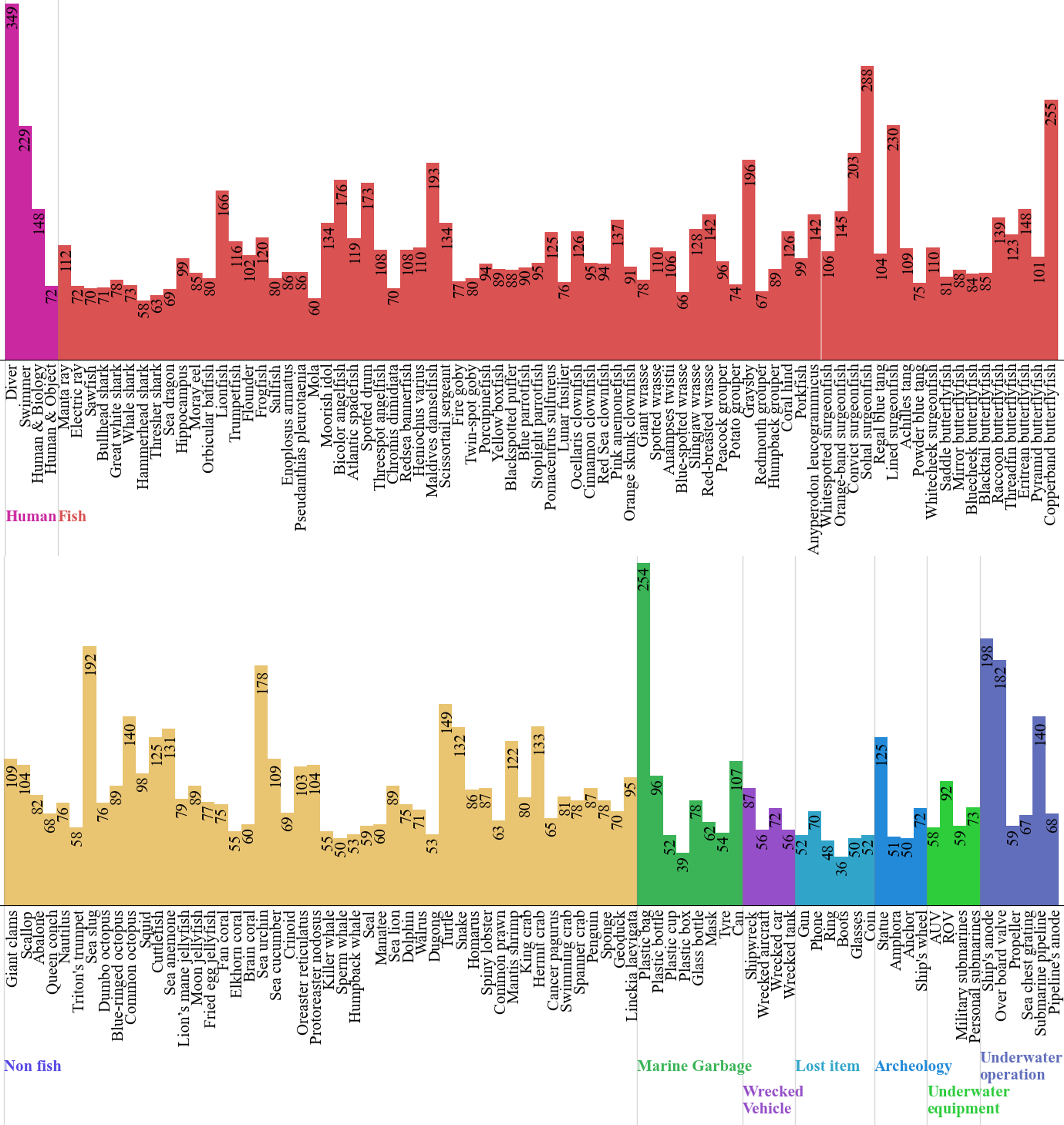}
\end{center}
   \caption{Number of underwater images for each salient object category. Best viewed on screen and zoomed-in for details.}
\label{fig:category}
\end{figure}

The construction of the USIS16K dataset emphasized diversity and representativeness of object categories by including 158 common underwater object categories found in natural underwater environments. A hierarchical taxonomic system was adopted to systematically organize these objects, as shown in Figure~\ref{fig:tax_sys}. 
The identification of object categories in the USIS16K was guided by a comprehensive literature review spanning the fields of 
underwater infrastructure inspection, underwater environmental monitoring, marine archaeology, marine biology, underwater resource exploration, and underwater robotic operation.
Based on these findings, 11 super-categories were defined: obstacles, facilities, underwater animals, humans, relics, marine fishes, plants, and litter, as shown in Table~\ref{tab:super-categories}. Each super-category was further subdivided into sub-categories based on internal attributes. The USIS16K dataset includes 158 distinct object categories commonly encountered in underwater environments.
Figure~\ref{fig:category} shows the distribution of images across object categories, with the number of images per category ranging from 36 to 258.

\begin{table}[]
\begin{center}
\caption{Descriptions of super-category in the USIS16K dataset.}
\label{tab:super-categories}
\resizebox{\linewidth}{!}{ 
\begin{tabular}{l|l}
\toprule
 \rowcolor[gray]{0.7}    Super-Category          & Descriptions   \\ \hline   
       Human participant     & Image content involving humans                             \\
  \rowcolor[gray]{0.9}     Fish                  & Common fish in underwater ecosystems                       \\ 
       Shellfish             & Common Shellfish in the underwater world                   \\
 \rowcolor[gray]{0.9}      Marine animals        & Common organisms living underwater                         \\
       Underwater trash      & Common trash in the underwater world                       \\     
 \rowcolor[gray]{0.9}      Underwater vehicles   & Underwater vehicles for various purposes                   \\
       Lost items            & Lost items by humans                                       \\
 \rowcolor[gray]{0.9}      Underwater facilities & Structures or installations built to operate underwater    \\
       Underwater relics     & Historical artifacts and structures found underwater       \\
 \rowcolor[gray]{0.9}      Underwater wrecks     & Remains of flights and vehicles found underwater           \\
       Others                & Other common underwater objects                            \\
       
\bottomrule
\end{tabular}}
\end{center}
\end{table}

\subsubsection{Underwater Scenes}
\begin{figure}[htbp]
\begin{center}
\includegraphics[width=0.99\linewidth]{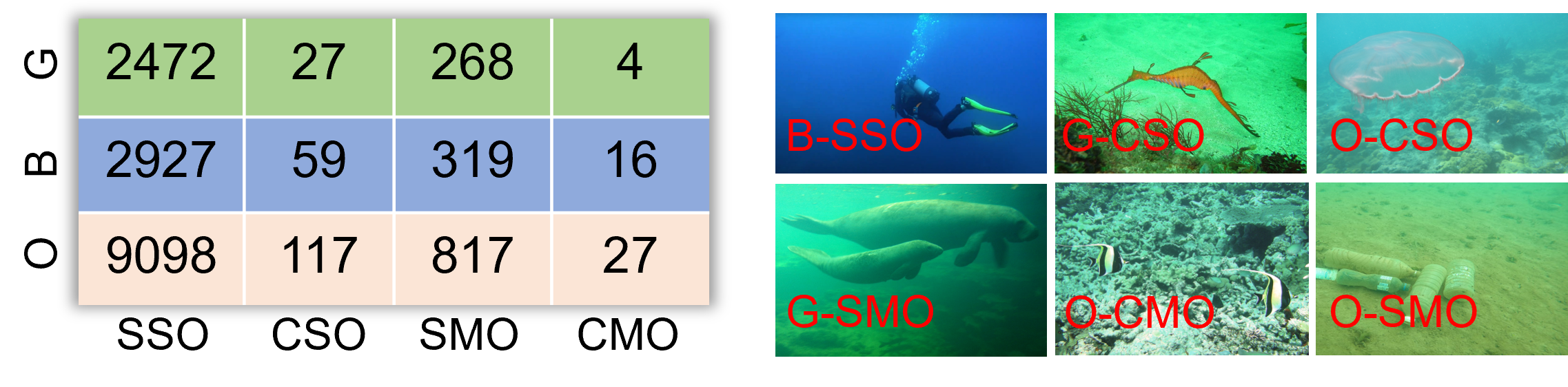}
\end{center}
   \caption{Attribute distribution of the images in the USIS16K.}
\label{fig:attribute}
\end{figure}

To comprehensively represent the diversity of underwater scenes in the USIS16K dataset, images were first classified based on background complexity as simple (S) or complex (C), and then categorized by the number of objects as single object (SO) or multiple objects (MO).
Additionally, the style of the underwater images are categorization based on light attenuation effects into greenish (G), blueish (B), or others (O). 
The overall distribution of these attributes in the USIS16K dataset is presented in Figure~\ref{fig:attribute}.

\subsection{Dataset Characteristics}
To provide deeper insights into the USIS16K dataset, a comprehensive statistical analysis was conducted to unveil its essential characteristics.

\subsubsection{Object Size}
\begin{figure}[ht]
\begin{center}
\includegraphics[width=0.99\linewidth]{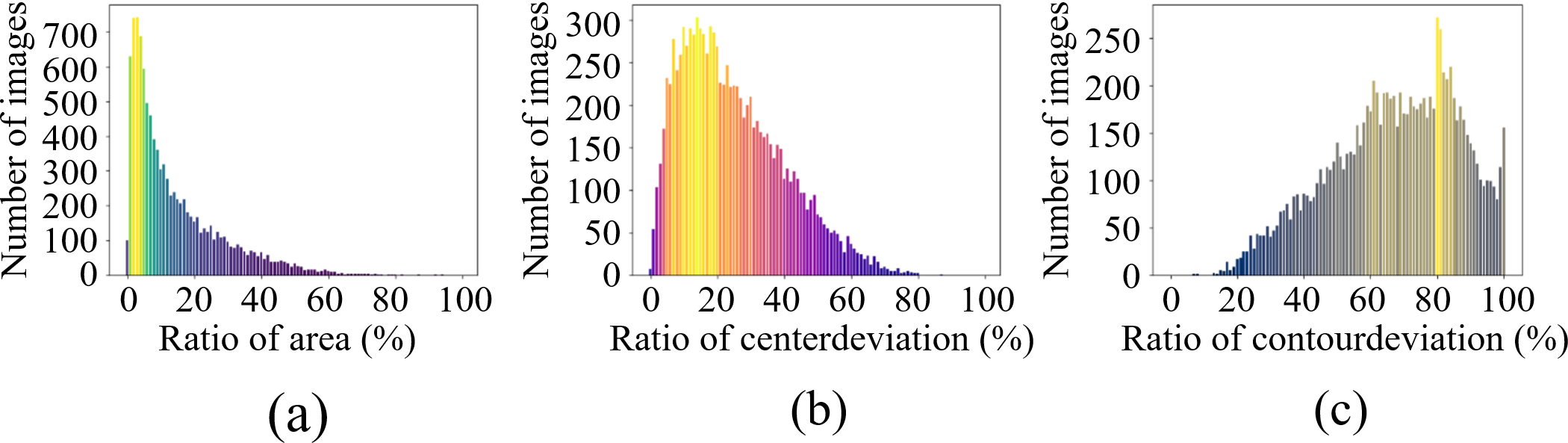}
\end{center}
   \caption{Essential characteristics of the USIS16K dataset. (a) Size of salient objects. (b) Center deviation. (c) Edge deviation.}
\label{fig:c1}
\end{figure}

The object size in the USIS16K dataset, represented by the area of the object, is defined as the number of pixels it occupies. 
To account for varying image resolutions, the calculated area is normalized by the total number of pixels in the respective image~\cite{6909437}. 
All objects in the dataset fall within the size range of [0.05\%, 93.98\%], as shown in Figure~\ref{fig:c1} (a), with an average size of 14.12\%, illustrating a broad spectrum of object sizes.
For further analysis, the calculated object sizes were categorized into three groups: Large:(Object size \textgreater 30\%), Middle: (5\% \textless Object size \textless 30\%) Small:(Object size \textless 5\%), as detailed in Table~\ref{tab:dataset_instance}. 
The corresponding statistics of the number of images in these size groups are 2676, 11078, and 2397, respectively, following a size distribution ratio of 1:4:1.

\subsubsection{Object Number}
The objects in the USIS16K dataset are pixel-wise annotated.
Considering that instance segmentation~\cite{hafiz2020survey} is a fast-developing branch in computer vision, the USIS16K dataset also contains instance segmentation annotations. As shown in Table~\ref{tab:dataset_instance},  
the numbers of images with 1, 2, and $\geq 3$ objects are 14700, 1228, and 223, respectively. 
\begin{table}
\small
\begin{center}
 \caption{Statistics of the object size and numbers in the proposed USIS16K dataset.
}
\begin{tabular}{l||ccc|ccc}
\toprule
  \multirow{2}{*}{\textbf{USIS16K}}
  &\multicolumn{3}{c}{Object Size}  &\multicolumn{3}{c}{Object Number}
 \\
& Large  &Middle &Small &1 &  2 & $\geq 3$ \\
\hline
  \rowcolor[gray]{0.9}  Images   &2676  &11078  &2397  &14700  &1228  &223\\
\toprule
\end{tabular}
\label{tab:dataset_instance}
\end{center}
\end{table}

\subsubsection{Object Location}
To show the location of an object within an image in the USIS16K dataset, two types of deviations are considered: center deviation (the distance from the center of the object to the center of the image) and edge deviation (the maximum distance from any point on the object to the center of the image). Following the method presented in \cite{fan2018SOC}, these distances are normalized by dividing by half the diagonal length of the image. The distribution of object locations in the USIS16K dataset is shown in Figure~\ref{fig:c1} (b) and (c), which shows a slight center bias among the objects.
\subsubsection{Channel Intensity}
\begin{figure}[ht]
\begin{center}
\includegraphics[width=0.99\linewidth]{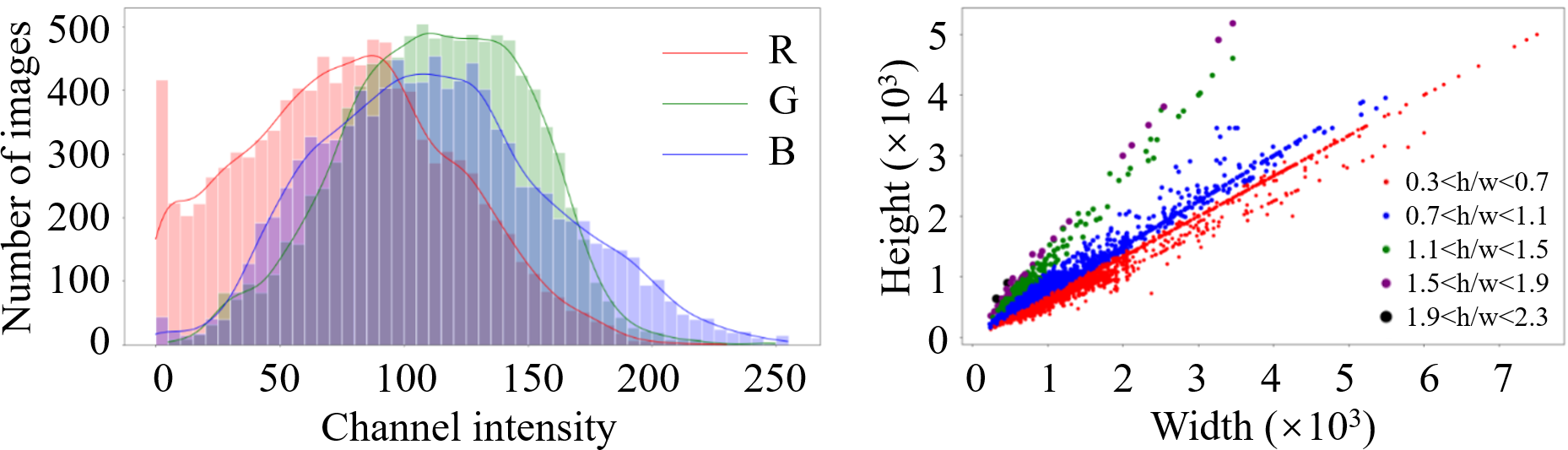}
\end{center}
   \caption{Essential characteristics of the USIS16K dataset. (a) Channel
intensity. (e) Resolution distribution.}
\label{fig:c2}
\end{figure}

Underwater images inevitably suffer from color attenuation due to the scattering and absorption of light as it travels through the water medium. Typically, the attenuation of the red channel is significantly greater than that of the blue and green channels~\cite{1994Light}. As a result, underwater images often exhibit a bluish or greenish appearance. 
To quantify this color attenuation characteristic in the USIS16K dataset, the average channel intensity of R, G, and B was calculated for each image.
The intensity distribution of R, G, and B channels across all images in the USIS16K dataset is shown in Figure~\ref{fig:c2} (a), where it is evident that the R channel consistently has the lowest intensity. 
\subsubsection{Resolution Distribution}
Images with varying resolutions can capture different visual scales and levels of detail. By incorporating such images in the USIS16K dataset, the model can learn object features across diverse conditions (e.g., distance, clarity) during model training, which enhances its generalization ability and reduces the risk of overfitting. Figure~\ref{fig:c2} (b) illustrates the resolution distribution of the USIS16K dataset, showing pixel counts in both the width and height directions. The width ranges from 225 to 7500 pixels, while the height spans from 135 to 5184 pixels. This distribution clearly demonstrates that the USIS16K dataset is capable of supporting high-quality model training.

%% file: 3_benchmark.tex
\section{Benchmark Evaluation}
\subsection{Underwater Object Detection}
Based on the established USIS16K dataset, we conduct a comprehensive evaluation of 8 object detection methods: SSD~\cite{liu2016ssd}, RetinaNet~\cite{lin2017focal}, ATSS~\cite{zhang2020bridging}, YOLOF~\cite{chen2021you}, Faster R-CNN~\cite{Ren_2017}, Cascade R-CNN~\cite{cai2018cascade}, Libra R-CNN~\cite{pang2019libra}, and Dynamic R-CNN~\cite{zhang2020dynamic}.
To ensure a fair comparison, all 8 models are retrained on the USIS16K dataset using their respective default configurations.
The quantitative performance of these detectors are evaluated by the commonly used evaluation metrics including frame per second (FPS), $mAP$, $AP_{50}$, $AP_{75}$, $AP_{M}$, $AP_{L}$, and model parameters (Params), the results are summarized in Table~\ref{tab:competitors1}.
\begin{table}[]
\begin{center}
\caption{Quantitative evaluation results of 8 object detection methods. The top-3 results are marked in \textcolor{red}{red}, \textcolor{blue}{blue}, and \textcolor{cyan}{cyan}.}
\resizebox{\linewidth}{!}{
\begin{tabular}{r|c|c|ccccc|c}
\toprule
\rowcolor[gray]{0.7} \textbf{Methods} & Backbone & FPS $(\uparrow)$ & $mAP (\uparrow)$ & $AP_{50}  (\uparrow)$ & $AP_{75} (\uparrow)$  & $AP_{M} (\uparrow)$ & $AP_{L}(\uparrow)$ & Params ($\downarrow$) \\
\hline
\hline
 SSD~\cite{he2017mask}                    & VGG16  & \textcolor{red}{60.4} & 69.5 & 88.2 & 79.2 & 47.9 & 70.1 & \textcolor{cyan}{44.7M}  \\
 
\rowcolor[gray]{0.9} RetinaNet~\cite{huang2019mask} & RXt101  &8.2 & 67.4  &77.7  & 72.3  &\textcolor{blue}{59.3} & 68.0 &  58.2M\\  

ATSS~\cite{chen2019hybrid}     &R50  & \textcolor{cyan}{17} & 69.0 & 79.5 & 75 & 56.6 & 69.6  & \textcolor{red}{32.5M}\\

\rowcolor[gray]{0.9} YOLOF~\cite{bolya2019yolact}   &R50 & \textcolor{blue}{26.1} & \textcolor{blue}{73.7} & 89.4 & \textcolor{cyan}{82.4} & \textcolor{cyan}{59.1} & \textcolor{blue}{74.3} & 46.0M \\ 

Faster R-CNN~\cite{wang2020solo}     & RXt101 &  8.6 & \textcolor{cyan}{72.1} & \textcolor{red}{91.3} & \textcolor{blue}{83.5} & 55.8 & \textcolor{cyan}{72.7} & 60.8M\\ 

\rowcolor[gray]{0.9} Cascade R-CNN~\cite{wang2020solov2} & RXt101 & 6.9 & \textcolor{red}{79.0} & \textcolor{blue}{90.7} & \textcolor{red}{86.2} & \textcolor{red}{61.2} & \textcolor{red}{79.7} & 88.3M \\   

Libra R-CNN~\cite{kirillov2020pointrend} & R101 & 10.7 & 70.1 & 89.4 & 80.3 & 57.2 & 70.7 & 61.4M \\ 

\rowcolor[gray]{0.9}Dynamic R-CNN~\cite{qiao2021detectors}  & R50  & 14.1 & 68.8 & \textcolor{cyan}{90.0} & 81.7 & 56.6 & 69.4 & \textcolor{blue}{42.2M}\\     
\toprule
\end{tabular}}
\label{tab:competitors1}
\end{center}
\end{table}

\begin{figure}[ht]
\begin{center}
\includegraphics[width=0.99\linewidth]{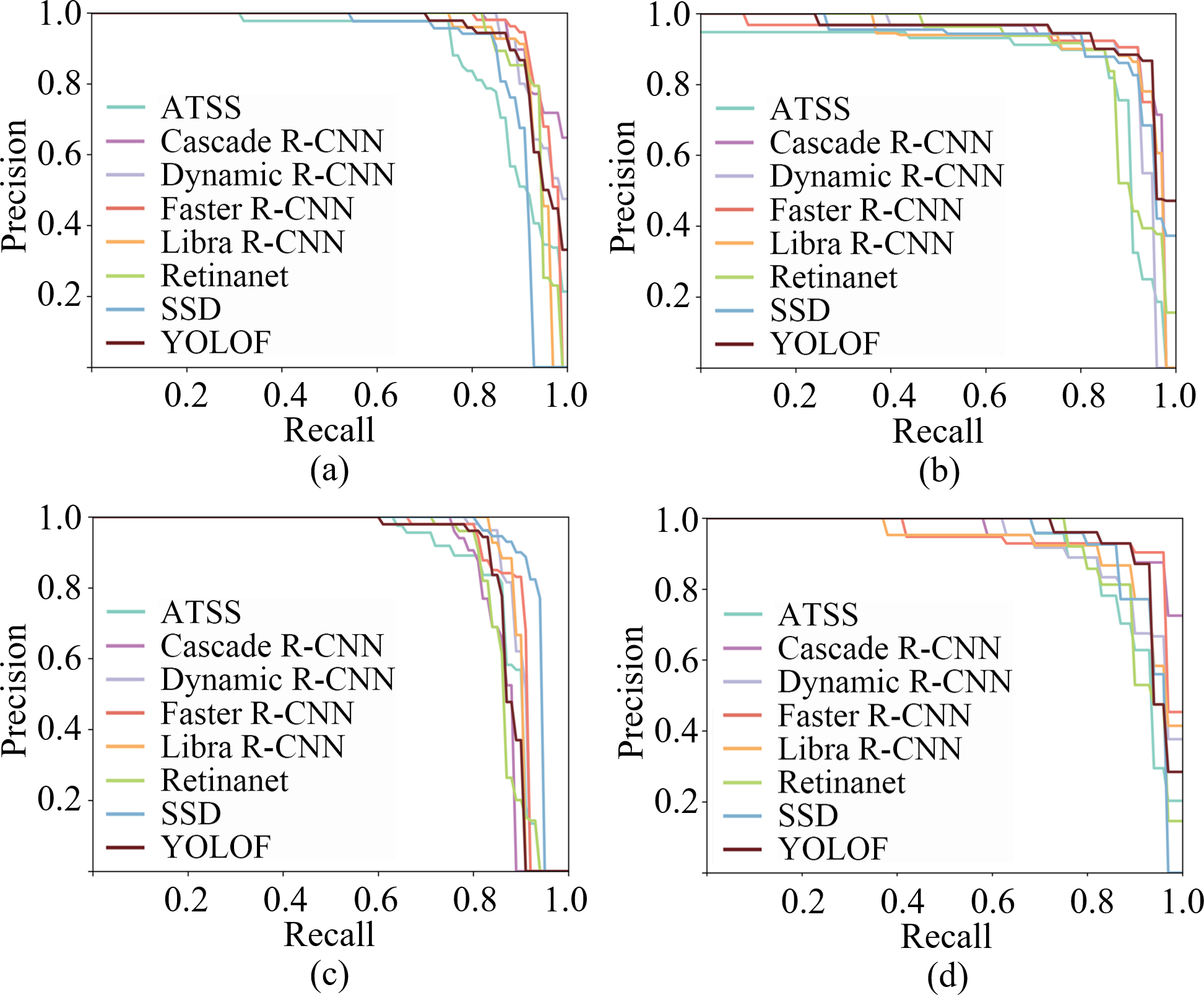}
\end{center}
\caption{PR curve comparison of 8 object detection methods evaluated on four representative object categories (Diver, Redmouth grouper, Regal blue tang, and Spiny lobster) from the USIS16K.}
\label{fig:PR}
\end{figure}

Cascade R-CNN outperforms all other seven methods across four key metrics: $mAP$, $AP_{75}$, $AP_{M}$, and $AP_{L}$.
In terms of FPS, the top-three methods are SSD, YOLOF, and ATSS, indicating their suitability for real-time applications.
For $AP_{50}$, the leading approaches are Faster R-CNN, Cascade R-CNN, and Dynamic R-CNN.
When comparing parameter counts (Params), the most lightweight (Top-3) models are ATSS, Dynamic R-CNN, and YOLOF. Figure~\ref{fig:PR} presents the precision-recall (PR) curves, showing that Cascade R-CNN outperforms most methods, though there remains room for improvement in specific precision-recall regions.
It is worth noting that SSD, RetinaNet, ATSS, and YOLOF are single-stage object detectors, whereas Faster R-CNN, Cascade R-CNN, Libra R-CNN, and Dynamic R-CNN are two-stage detectors.
Single-stage object detectors are generally more lightweight and faster, making them suitable for resource-constrained or real-time scenarios. However, they often lag behind two-stage detectors in terms of detection accuracy, particularly for small or occluded objects.

\subsection{Underwater Salient Instance Segmentation}
\begin{figure}[ht]
\begin{center}
\includegraphics[width=0.99\linewidth]{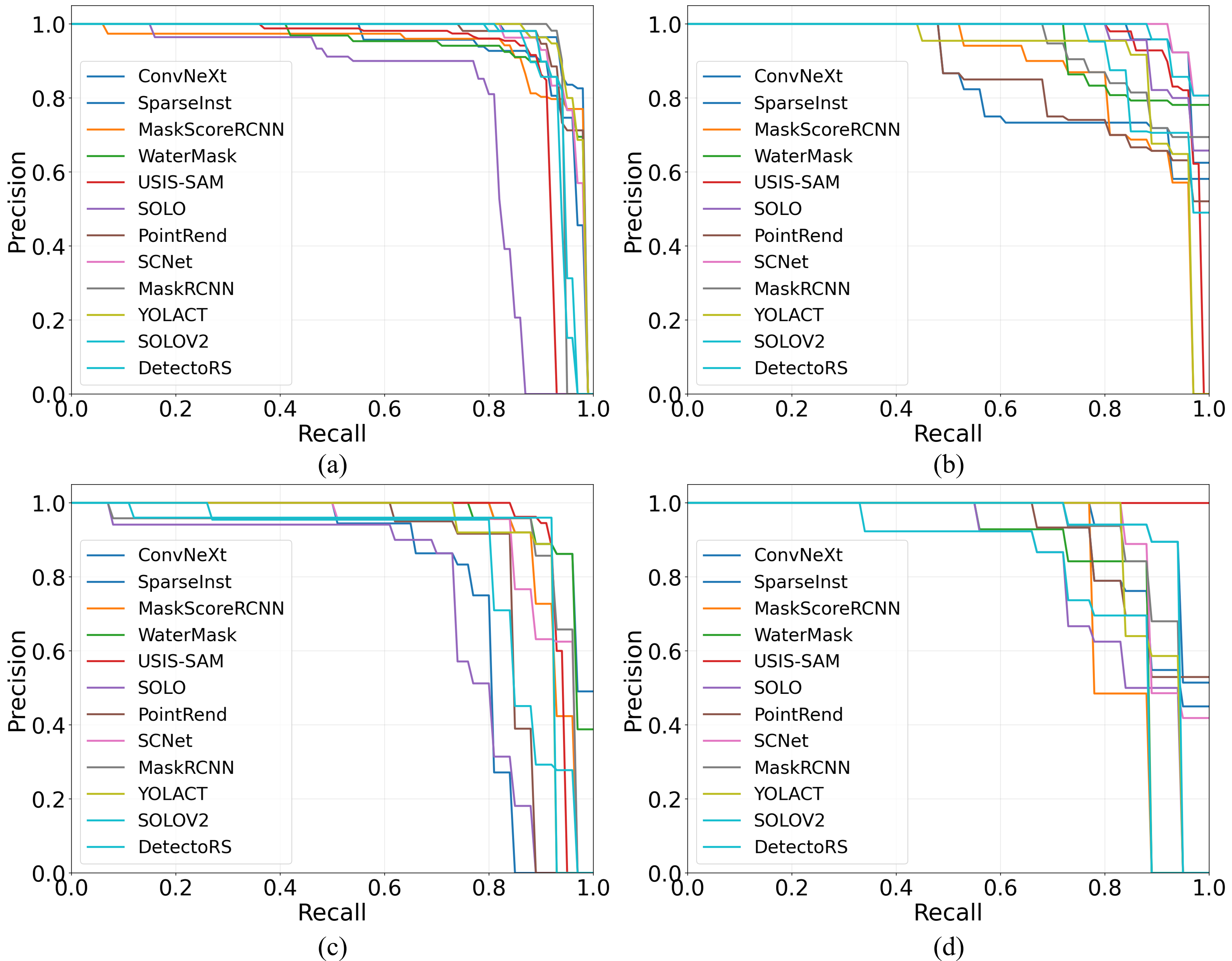}
\end{center}
\caption{PR curve comparison of 12 instance segmentation methods evaluated on four representative object categories (Diver, Swimmer, Plastic cup, and Sea chest grating) from the USIS16K.}
\label{fig:PR2}
\end{figure}

Furthermore, based on the USIS16K dataset, we conduct a comprehensive evaluation of 12 instance segmentation methods: 10 designed originally for terrestrial environments, including Mask R-CNN \cite{he2017mask}, ConvNeXt \cite{liu2022convnet}, Mask Scoring R-CNN \cite{huang2019mask}, YOLACT \cite{bolya2019yolact}, SOLO \cite{wang2020solo}, SOLOv2 \cite{wang2020solov2}, PointRend \cite{kirillov2020pointrend}, DetectoRS \cite{qiao2021detectors}, SCNet \cite{vu2021scnet}, and SparseInst \cite{Cheng2022SparseInst}, one underwater semantic-segmentation model (WaterMask \cite{lian2023watermask}), and one dedicated USIS approach (USIS-SAM \cite{lian2024diving}).
The quantitative performance of these detectors are evaluated by the commonly used evaluation metrics, including $mAP$, $AP_{50}$, $AP_{75}$, $AP_{S}$, $AP_{M}$, $AP_{L}$, and Params, the results are summarized in Table~\ref{tab:competitors2}.
ConvNeXt leads on four metrics: $AP_{50}$, $AP_{75}$, $AP_{S}$, and $AP_{M}$. For overall $mAP$, the top-3 methods are USIS-SAM, ConvNeXt, and PointRend; for $AP_{L}$, the leaders are USIS-SAM, ConvNeXt, and DetectoRS; and for model size (Params), DetectoRS, SparseInst, and YOLACT have the fewest parameters. Figure \ref{fig:PR2} displays the PR curve comparison of 12 methods evaluated on the randomly selected four categories of objects in USIS16K dataset. It can be seen that YOLOCAT outperforms most other methods in terms of PR curve performance on three categories, Diver, Swimmer, and Plastic Cup, although there remains room for further improvement. USIS-SAM and WaterMask, as underwater-specific models, outperform most methods developed for terrestrial environments, demonstrating the effectiveness of domain-specific designs.
\begin{table}[]
\begin{center}
\caption{Quantitative evaluation results of 12 instance segmentation methods. The top-3 results are marked in \textcolor{red}{red}, \textcolor{blue}{blue}, and \textcolor{cyan}{cyan}.}
\resizebox{\linewidth}{!}{
\begin{tabular}{r||c|cc|ccc|c}
\toprule
\rowcolor[gray]{0.7} \textbf{Methods}  & $mAP (\uparrow)$ & $AP_{50}  (\uparrow)$ & $AP_{75} (\uparrow)$ 
& $AP_{S} (\uparrow) $ & $AP_{M} (\uparrow)$ & $AP_{L}  (\uparrow)$ & Params ($\downarrow$)  \\
\hline
\hline
Mask R-CNN~\cite{he2017mask}            & .736  &.900 & .817
                            &.400  & .604  &.745 & 61.83M \\
\rowcolor[gray]{0.9} Mask Scoring R-CNN~\cite{huang2019mask} & .716  &.885 & .798 
                            &.450  & .515  &.726 & 61.23M \\  
YOLACT~\cite{bolya2019yolact}     & .754  &.904 & .826 
                            &.450  & .545 &.760  & \textcolor{cyan}{35.86M} \\  
\rowcolor[gray]{0.9}SOLO~\cite{wang2020solo}     & .595  &.763 & .659 
                            &.119  & .327  &.606 & 36.48M\\ 
 SOLOV2~\cite{wang2020solov2} & .717  &.859 & .771
                            &.450  & .482  &.726 & 45.96M \\   
\rowcolor[gray]{0.9}PointRend~\cite{kirillov2020pointrend}  & \textcolor{cyan}{.763}  &.894 & .827 
                            &\textcolor{blue}{.500}  & \textcolor{cyan}{.659} &.774 & 64.69M \\   
DetectoRS~\cite{qiao2021detectors}  & .732  &\textcolor{blue}{.938} & \textcolor{blue}{.872} 
                            &.433  & \textcolor{blue}{.664}  &\textcolor{cyan}{.788} & \textcolor{red}{13.60M}\\     
\rowcolor[gray]{0.9}SCNet~\cite{vu2021scnet}    & .752  &\textcolor{cyan}{.910} & .822 
                            &\textcolor{cyan}{. 475} & .630  &.759 & 90.10M\\
SparseInst~\cite{Cheng2022SparseInst}   & .730  &.850 & .769 
                            &.450  & .449  &.740 & \textcolor{blue}{31.65M} \\
 \rowcolor[gray]{0.9}ConvNeXt~\cite{liu2022convnet}     & \textcolor{blue}{.785}  & \textcolor{red}{.953} & \textcolor{red}{.879} 
                            &\textcolor{red}{.551}  & \textcolor{red}{.688}  &\textcolor{blue}{.792}  & {48.52M}  \\  
Watermask~\cite{lian2023watermask}          & .727  &.868 & .793 
                            &.417 & .570  &{.736} & {48.27M}  \\ 
\rowcolor[gray]{0.9}USIS-SAM~\cite{lian2024diving}  & \textcolor{red}{.810}  &.908 & \textcolor{cyan}{.871} 
                            &.450  & .643  &\textcolor{red}{.818} & 698.9M  \\ 
\toprule
\end{tabular}}
\label{tab:competitors2}
\end{center}
\end{table}

\begin{figure*}[ht]
\begin{center}
\includegraphics[width=1\linewidth]{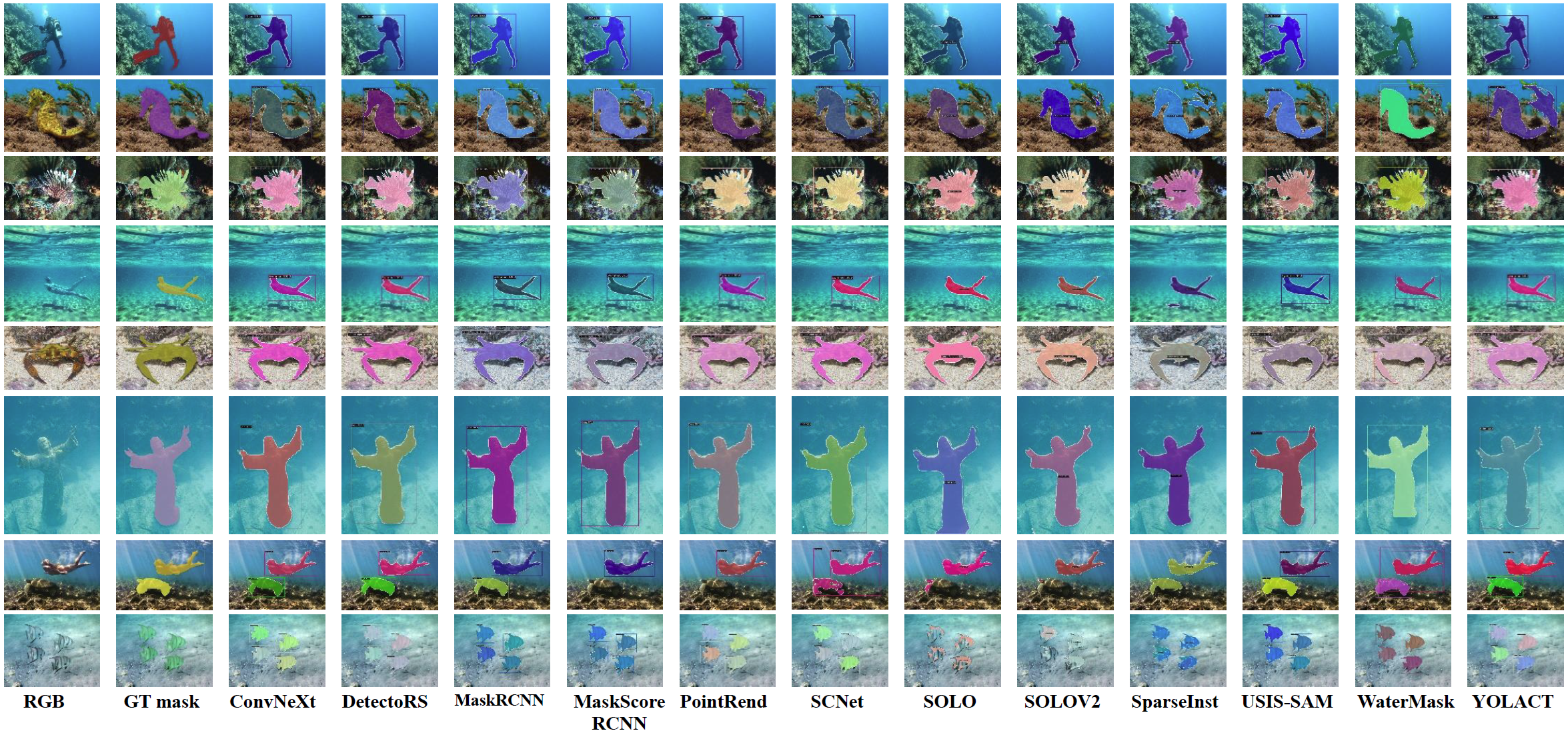}
\end{center}
    \caption{Qualitative comparison of 12 instance segmentation methods on the USIS16K dataset.}
\label{fig:qualitative_com}
\end{figure*}

Figure~\ref{fig:qualitative_com} presents qualitative comparisons of the 12 methods by visualizing 8 randomly selected images from the test set. ConvNeXt performs robustly across diverse scenarios, including single- and multiple-instance scenes, objects of complex and simple shapes, and categories such as divers, animals, and plants. For the test images containing a single object, all methods achieve accurate USIS results, except for the second row image. In this case, which features a sea dragon, all methods successfully detect the object; however, Mask Scoring R-CNN, PointRend, SOLOv2, and SparseInst exhibit imprecise boundary predictions. Meanwhile, DetectoRS and SCNet, although capable of correctly localizing the object, fail to produce fine-grained segmentation. For the image containing two objects and with a complex background (second-to-last row), Mask Scoring R-CNN, PointRend, SOLO, and SOLOv2 miss at least one salient object. In the last row, where the test image contains four objects against a simple background, all 12 methods successfully detect each salient object and produce accurate instance segmentations.

%% file: 4_conclusion.tex
\section{Conclusion}
In this paper, we present USIS16K, a large-scale, high-quality dataset designed to advance research in USIS. The dataset consists of 16,151 underwater images, covering 158 object categories commonly found in marine environments. Each image is annotated with instance-level masks for the accurately identified salient objects, enabling a wide range of vision tasks such as underwater object detection and USIS.
To support further research, we provide benchmark evaluations for both underwater object detection and USIS tasks on USIS16K, along with a comprehensive comparative analysis. We argue that USIS16K and its accompanying benchmarks are a valuable resource for the research community, fostering progress in this emerging area. Despite recent advancements, USIS remains a challenging and largely unsolved problem, offering substantial opportunities for future exploration.

%% file: acknowledge.tex
\section{Acknowledgement}
\label{Sec7}
We would like to express our gratitude to all those who assisted us in the construction of the dataset. This work is supported by the Shenzhen Science and Technology Program  (JSGG20211029095205007) and the Shenzhen Science and Technology Major Program (KJZD20231023100459001).

%% file: main.bbl
\begin{thebibliography}{10}

\bibitem{akkaynak2018revised}
Derya Akkaynak and Tali Treibitz.
\newblock A revised underwater image formation model.
\newblock In {\em Proceedings of the IEEE conference on computer vision and pattern recognition}, pages 6723--6732, 2018.

\bibitem{bolya2019yolact}
Daniel Bolya, Chong Zhou, Fanyi Xiao, and Yong~Jae Lee.
\newblock Yolact: Real-time instance segmentation.
\newblock In {\em Proceedings of the IEEE/CVF international conference on computer vision}, pages 9157--9166, 2019.

\bibitem{cai2018cascade}
Zhaowei Cai and Nuno Vasconcelos.
\newblock Cascade r-cnn: Delving into high quality object detection.
\newblock In {\em Proceedings of the IEEE conference on computer vision and pattern recognition}, pages 6154--6162, 2018.

\bibitem{chen2019hybrid}
Kai Chen, Jiangmiao Pang, Jiaqi Wang, Yu~Xiong, Xiaoxiao Li, Shuyang Sun, Wansen Feng, Ziwei Liu, Jianping Shi, Wanli Ouyang, et~al.
\newblock Hybrid task cascade for instance segmentation.
\newblock In {\em Proceedings of the IEEE/CVF conference on computer vision and pattern recognition}, pages 4974--4983, 2019.

\bibitem{chen2021you}
Qiang Chen, Yingming Wang, Tong Yang, Xiangyu Zhang, Jian Cheng, and Jian Sun.
\newblock You only look one-level feature.
\newblock In {\em IEEE Conference on Computer Vision and Pattern Recognition}, 2021.

\bibitem{Cheng2022SparseInst}
Tianheng Cheng, Xinggang Wang, Shaoyu Chen, Wenqiang Zhang, Qian Zhang, Chang Huang, Zhaoxiang Zhang, and Wenyu Liu.
\newblock Sparse instance activation for real-time instance segmentation.
\newblock In {\em Proc. IEEE Conf. Computer Vision and Pattern Recognition (CVPR)}, 2022.

\bibitem{deng2009imagenet}
Jia Deng, Wei Dong, Richard Socher, Li-Jia Li, Kai Li, and Li~Fei-Fei.
\newblock Imagenet: A large-scale hierarchical image database.
\newblock In {\em 2009 IEEE conference on computer vision and pattern recognition}, pages 248--255. Ieee, 2009.

\bibitem{fan2018salient}
Deng-Ping Fan, Ming-Ming Cheng, Jiang-Jiang Liu, Shang-Hua Gao, Qibin Hou, and Ali Borji.
\newblock Salient objects in clutter: Bringing salient object detection to the foreground.
\newblock In {\em Proceedings of the European conference on computer vision (ECCV)}, pages 186--202, 2018.

\bibitem{fan2018SOC}
Deng-Ping Fan, Ming-Ming Cheng, Jiang-Jiang Liu, Shang-Hua Gao, Qibin Hou, and Ali Borji.
\newblock Salient objects in clutter: Bringing salient object detection to the foreground.
\newblock In {\em European Conference on Computer Vision (ECCV)}. Springer, 2018.

\bibitem{hafiz2020survey}
Abdul~Mueed Hafiz and Ghulam~Mohiuddin Bhat.
\newblock A survey on instance segmentation: state of the art.
\newblock {\em International journal of multimedia information retrieval}, 9(3):171--189, 2020.

\bibitem{he2017mask}
Kaiming He, Georgia Gkioxari, Piotr Doll{\'a}r, and Ross Girshick.
\newblock Mask r-cnn.
\newblock In {\em Proceedings of the IEEE international conference on computer vision}, pages 2961--2969, 2017.

\bibitem{10102831}
Lin Hong, Xin Wang, Gan Zhang, and Ming Zhao.
\newblock Usod10k: A new benchmark dataset for underwater salient object detection.
\newblock {\em IEEE Transactions on Image Processing}, pages 1--1, 2023.

\bibitem{huang2019mask}
Zhaojin Huang, Lichao Huang, Yongchao Gong, Chang Huang, and Xinggang Wang.
\newblock Mask scoring r-cnn.
\newblock In {\em Proceedings of the IEEE/CVF conference on computer vision and pattern recognition}, pages 6409--6418, 2019.

\bibitem{9340821}
Md~Jahidul Islam, Chelsey Edge, Yuyang Xiao, Peigen Luo, Muntaqim Mehtaz, Christopher Morse, Sadman~Sakib Enan, and Junaed Sattar.
\newblock Semantic segmentation of underwater imagery: Dataset and benchmark.
\newblock In {\em 2020 IEEE/RSJ International Conference on Intelligent Robots and Systems (IROS)}, pages 1769--1776, 2020.

\bibitem{Islam-RSS-22}
{Md Jahidul} Islam, Ruobing Wang, and Junaed Sattar.
\newblock {SVAM: Saliency-guided Visual Attention Modeling by Autonomous Underwater Robot}.
\newblock In {\em Proceedings of Robotics: Science and Systems}, New York City, NY, USA, June 2022.

\bibitem{kirillov2020pointrend}
Alexander Kirillov, Yuxin Wu, Kaiming He, and Ross Girshick.
\newblock Pointrend: Image segmentation as rendering.
\newblock In {\em Proceedings of the IEEE/CVF conference on computer vision and pattern recognition}, pages 9799--9808, 2020.

\bibitem{li2017instance}
Guanbin Li, Yuan Xie, Liang Lin, and Yizhou Yu.
\newblock Instance-level salient object segmentation.
\newblock In {\em Proceedings of the IEEE conference on computer vision and pattern recognition}, pages 2386--2395, 2017.

\bibitem{li2025uwsam}
Hua Li, Shijie Lian, Zhiyuan Li, Runmin Cong, and Sam Kwong.
\newblock Uwsam: Segment anything model guided underwater instance segmentation and a large-scale benchmark dataset.
\newblock {\em arXiv preprint arXiv:2505.15581}, 2025.

\bibitem{li2021ion}
Weijie Li, Xinhang Song, Yubing Bai, Sixian Zhang, and Shuqiang Jiang.
\newblock Ion: Instance-level object navigation.
\newblock In {\em Proceedings of the 29th ACM international conference on multimedia}, pages 4343--4352, 2021.

\bibitem{6909437}
Yin Li, Xiaodi Hou, Christof Koch, James~M. Rehg, and Alan~L. Yuille.
\newblock The secrets of salient object segmentation.
\newblock In {\em 2014 IEEE Conference on Computer Vision and Pattern Recognition}, pages 280--287, 2014.

\bibitem{lian2023watermask}
Shijie Lian, Hua Li, Runmin Cong, Suqi Li, Wei Zhang, and Sam Kwong.
\newblock Watermask: Instance segmentation for underwater imagery.
\newblock In {\em Proceedings of the IEEE/CVF International Conference on Computer Vision}, pages 1305--1315, 2023.

\bibitem{lian2024diving}
Shijie Lian, Ziyi Zhang, Hua Li, Wenjie Li, Laurence~Tianruo Yang, Sam Kwong, and Runmin Cong.
\newblock Diving into underwater: Segment anything model guided underwater salient instance segmentation and a large-scale dataset.
\newblock {\em arXiv preprint arXiv:2406.06039}, 2024.

\bibitem{lin2017focal}
Tsung-Yi Lin, Priya Goyal, Ross Girshick, Kaiming He, and Piotr Doll{\'a}r.
\newblock Focal loss for dense object detection.
\newblock In {\em Proceedings of the IEEE international conference on computer vision}, 2017.

\bibitem{lin2014microsoft}
Tsung-Yi Lin, Michael Maire, Serge Belongie, James Hays, Pietro Perona, Deva Ramanan, Piotr Doll{\'a}r, and C~Lawrence Zitnick.
\newblock Microsoft coco: Common objects in context.
\newblock In {\em European conference on computer vision}, pages 740--755. Springer, 2014.

\bibitem{liu2021scg}
Nian Liu, Wangbo Zhao, Ling Shao, and Junwei Han.
\newblock Scg: Saliency and contour guided salient instance segmentation.
\newblock {\em IEEE Transactions on Image Processing}, 30:5862--5874, 2021.

\bibitem{liu2016ssd}
Wei Liu, Dragomir Anguelov, Dumitru Erhan, Christian Szegedy, Scott Reed, Cheng-Yang Fu, and Alexander~C Berg.
\newblock Ssd: Single shot multibox detector.
\newblock In {\em Computer Vision--ECCV 2016: 14th European Conference, Amsterdam, The Netherlands, October 11--14, 2016, Proceedings, Part I 14}, pages 21--37. Springer, 2016.

\bibitem{liu2022convnet}
Zhuang Liu, Hanzi Mao, Chao-Yuan Wu, Christoph Feichtenhofer, Trevor Darrell, and Saining Xie.
\newblock A convnet for the 2020s.
\newblock In {\em Proceedings of the IEEE/CVF conference on computer vision and pattern recognition}, pages 11976--11986, 2022.

\bibitem{1994Light}
C.~D. Mobley.
\newblock Light and water: Radiative transfer in natural waters.
\newblock {\em Academic Press}, 1994.

\bibitem{pang2019libra}
Jiangmiao Pang, Kai Chen, Jianping Shi, Huajun Feng, Wanli Ouyang, and Dahua Lin.
\newblock Libra r-cnn: Towards balanced learning for object detection.
\newblock In {\em Proceedings of the IEEE/CVF conference on computer vision and pattern recognition}, pages 821--830, 2019.

\bibitem{pei2022transformer}
Jialun Pei, Tianyang Cheng, He~Tang, and Chuanbo Chen.
\newblock Transformer-based efficient salient instance segmentation networks with orientative query.
\newblock {\em IEEE Transactions on Multimedia}, 25:1964--1978, 2022.

\bibitem{10614124}
Jialun Pei, Tao Jiang, He~Tang, Nian Liu, Yueming Jin, Deng-Ping Fan, and Pheng-Ann Heng.
\newblock Calibnet: Dual-branch cross-modal calibration for rgb-d salient instance segmentation.
\newblock {\em IEEE Transactions on Image Processing}, 33:4348--4362, 2024.

\bibitem{qiao2021detectors}
Siyuan Qiao, Liang-Chieh Chen, and Alan Yuille.
\newblock Detectors: Detecting objects with recursive feature pyramid and switchable atrous convolution.
\newblock In {\em Proceedings of the IEEE/CVF conference on computer vision and pattern recognition}, pages 10213--10224, 2021.

\bibitem{redmon2016you}
Joseph Redmon, Santosh Divvala, Ross Girshick, and Ali Farhadi.
\newblock You only look once: Unified, real-time object detection.
\newblock In {\em Proceedings of the IEEE conference on computer vision and pattern recognition}, pages 779--788, 2016.

\bibitem{Ren_2017}
Shaoqing Ren, Kaiming He, Ross Girshick, and Jian Sun.
\newblock Faster r-cnn: Towards real-time object detection with region proposal networks.
\newblock {\em IEEE Transactions on Pattern Analysis and Machine Intelligence}, Jun 2017.

\bibitem{vu2021scnet}
Thang Vu, Haeyong Kang, and Chang~D Yoo.
\newblock Scnet: Training inference sample consistency for instance segmentation.
\newblock In {\em Proceedings of the AAAI Conference on Artificial Intelligence}, volume~35, pages 2701--2709, 2021.

\bibitem{wang2020solo}
Xinlong Wang, Tao Kong, Chunhua Shen, Yuning Jiang, and Lei Li.
\newblock Solo: Segmenting objects by locations.
\newblock In {\em Computer Vision--ECCV 2020: 16th European Conference, Glasgow, UK, August 23--28, 2020, Proceedings, Part XVIII 16}, pages 649--665. Springer, 2020.

\bibitem{wang2020solov2}
Xinlong Wang, Rufeng Zhang, Tao Kong, Lei Li, and Chunhua Shen.
\newblock Solov2: Dynamic and fast instance segmentation.
\newblock {\em Advances in Neural information processing systems}, 33:17721--17732, 2020.

\bibitem{9998080}
Maryna Waszak, Alexandre Cardaillac, Brian Elvesæter, Frode Rødølen, and Martin Ludvigsen.
\newblock Semantic segmentation in underwater ship inspections: Benchmark and data set.
\newblock {\em IEEE Journal of Oceanic Engineering}, 48(2):462--473, 2023.

\bibitem{xie2021unseen}
Christopher Xie, Yu~Xiang, Arsalan Mousavian, and Dieter Fox.
\newblock Unseen object instance segmentation for robotic environments.
\newblock {\em IEEE Transactions on Robotics}, 37(5):1343--1359, 2021.

\bibitem{zhang2020dynamic}
Hongkai Zhang, Hong Chang, Bingpeng Ma, Naiyan Wang, and Xilin Chen.
\newblock Dynamic r-cnn: Towards high quality object detection via dynamic training.
\newblock In {\em Computer Vision--ECCV 2020: 16th European Conference, Glasgow, UK, August 23--28, 2020, Proceedings, Part XV 16}, pages 260--275. Springer, 2020.

\bibitem{zhang2020bridging}
Shifeng Zhang, Cheng Chi, Yongqiang Yao, Zhen Lei, and Stan~Z Li.
\newblock Bridging the gap between anchor-based and anchor-free detection via adaptive training sample selection.
\newblock In {\em Proceedings of the IEEE/CVF conference on computer vision and pattern recognition}, pages 9759--9768, 2020.

\bibitem{zheng2024coralscop}
Ziqiang Zheng, Haixin Liang, Binh-Son Hua, Yue~Him Wong, Put Ang, Apple Pui~Yi Chui, and Sai-Kit Yeung.
\newblock Coralscop: Segment any coral image on this planet.
\newblock In {\em Proceedings of the IEEE/CVF Conference on Computer Vision and Pattern Recognition}, pages 28170--28180, 2024.

\end{thebibliography}
